\documentclass[11pt]{article}

\usepackage[preprint]{acl}

\usepackage{times}
\usepackage{latexsym}
\usepackage[T1]{fontenc}
\usepackage[utf8]{inputenc}
\usepackage{microtype}
\usepackage{inconsolata}
\usepackage{graphicx}
\usepackage{amsmath}
\usepackage{booktabs}
\usepackage{xcolor}
\definecolor{promptbg}{gray}{0.95}
\newsavebox{\promptboxcontent}
\newenvironment{promptbox}{%
  \par\noindent\begingroup
  \setlength{\fboxsep}{5pt}%
  \begin{lrbox}{\promptboxcontent}%
  \begin{minipage}{\dimexpr\linewidth-2\fboxsep\relax}%
  \setlength{\parindent}{0pt}%
  \setlength{\parskip}{0.4\baselineskip}%
}{%
  \end{minipage}%
  \end{lrbox}%
  \colorbox{promptbg}{\usebox{\promptboxcontent}}%
  \par\endgroup
}
\newsavebox{\exampleboxcontent}
\newenvironment{examplebox}{%
  \par\noindent\begingroup
  \setlength{\fboxsep}{5pt}%
  \setlength{\fboxrule}{0.4pt}%
  \begin{lrbox}{\exampleboxcontent}%
  \begin{minipage}{\dimexpr\linewidth-2\fboxsep-2\fboxrule\relax}%
  \setlength{\parindent}{0pt}%
  \setlength{\parskip}{0.4\baselineskip}%
}{%
  \end{minipage}%
  \end{lrbox}%
  \fbox{\usebox{\exampleboxcontent}}%
  \par\endgroup
}
\makeatletter
\g@addto@macro{\UrlBreaks}{%
  \do\a\do\b\do\c\do\d\do\e\do\f\do\g\do\h\do\i\do\j\do\k\do\l\do\m
  \do\n\do\o\do\p\do\q\do\r\do\s\do\t\do\u\do\v\do\w\do\x\do\y\do\z
  \do\0\do\1\do\2\do\3\do\4\do\5\do\6\do\7\do\8\do\9}
\makeatother
\newcommand{\appref}[1]{\hyperref[#1]{Appendix~\ref*{#1}}}

\title{Understanding Interpretation Difficulty in Harmful Online Communication: Insights from Cybercrime Communities}

\author{
Tomohiro Okatsu,
Naoki Takada,
Yin Min Pa Pa,
Katsunari Yoshioka,
Tatsunori Mori\\
Yokohama National University, Japan\\
\texttt{\{mori-tomohiro-bm, takada-naoki-xs\}@ynu.jp}\\
\texttt{\{yinminn-papa-jp, yoshioka, tmori\}@ynu.ac.jp}
}

\begin{document}
\maketitle

\begin{abstract}
Harmful online communication often contains slang, coded terms, abbreviations, and community-specific expressions, which make messages difficult to interpret. This paper presents an exploratory study of interpretation difficulty in Discord chats related to cybercrime. We construct reference interpretations of purposefully selected difficult messages, which were reviewed by an expert. We then use them to evaluate human and large language model (LLM) interpretations under different context conditions. The results show that local context alone is often insufficient for humans, while external knowledge and extended conversational context substantially improve human interpretation. For LLMs, local context also improves interpretation, and the larger model performs better. We further conduct a qualitative error analysis and propose a preliminary classification of factors that make harmful chats difficult to interpret. These findings suggest that harmful-content analysis should treat interpretation as an evidence-integration problem, rather than as message-level classification alone.

Warning: This paper contains examples and discussions of harmful online content.
\end{abstract}

\section{Introduction}

Cybercrime and other forms of harmful behavior on online platforms have become serious global problems \citep{interpol-2026-asp-cyber-threat,lukosiute-etal-2026-global}. Platforms such as Discord are increasingly used not only for ordinary communication but also for harmful activities, including malware distribution, credential theft, phishing, doxxing, and illicit trading \citep{sophos-2021-discord-malware,zscaler-2021-discord-cdn,heslep-berge-2024-mapping,intel471-2024-discord-cybercrime}.

Prior work has studied automated methods for detecting harmful online behavior, including grooming, cyberbullying, and hate speech \citep{vogt-etal-2021-early,murnion-etal-2018-machine}. However, harmful intent is often not expressed directly. Instead, users may rely on slang, coded terms, abbreviations, and community-specific expressions \citep{magu-luo-2018-determining,yuan-etal-2018-reading,hughes-etal-2023-argot}. Such expressions make harmful messages difficult not only to detect, but also to interpret.

Recent large language models (LLMs) have shown strong abilities in analyzing informal and context-dependent expressions \citep{brown-etal-2020-language,ziems-etal-2024-large}, which may make them useful tools for supporting the interpretation of difficult online messages. Nevertheless, rapidly changing slang and domain-specific expressions remain challenging for LLMs \citep{sun-etal-2024-toward,mei-etal-2024-slang,zhang-etal-2024-dont-go}. It is still unclear what makes harmful messages difficult to interpret and whether humans and LLMs struggle with the same types of difficulty.

In this paper, we conduct an exploratory study of interpretation difficulty in cybercrime-related Discord chats. Using difficult messages, we compare human and LLM interpretations under various context conditions and analyze recurring sources of interpretation failure. Based on this analysis, we propose a preliminary classification of factors that make harmful chats difficult to interpret.

\noindent The research questions of this paper are as follows:
\begin{itemize}
\setlength{\itemsep}{0pt}
\setlength{\parskip}{0pt}
\setlength{\parsep}{0pt}
\item \textbf{RQ1}: How do humans and LLMs interpret difficult harmful messages under different context conditions?
\item \textbf{RQ2}: What factors make harmful chats difficult to interpret?
\end{itemize}

\noindent Our contribution can be summarized as follows:

\begin{itemize}
\setlength{\itemsep}{0pt}
\setlength{\parskip}{0pt}
\setlength{\parsep}{0pt}
\item We evaluate human and LLM interpretations of difficult cybercrime-related Discord messages under various context conditions.
\item We classify recurring factors that make harmful chats difficult for humans and LLMs to interpret through qualitative error analysis.
\end{itemize}

\section{Related Work}

\subsection{Crime and Harmful Communication on Online Platforms}

Harmful and criminal online activity increasingly occurs on mainstream communication platforms, not only on underground forums or dark Web communities. Prior work has reported cybercrime-related activity on platforms such as Telegram and other widely used communication services, including malware distribution, phishing, scams, illicit trading, and coordinated abuse \citep{roy-etal-2025-darkgram,la-morgia-etal-2021-uncovering,europol-2024-encrypted,associated-press-2025-whatsapp,he-etal-2026-six-million,roy-etal-2021-evaluating,acharya-etal-2025-pirates,roy-etal-2024-unveiling}.

Discord is particularly relevant for studying such activity because it combines semi-closed communities with messaging, bots, file sharing, and voice chat. Prior reports and studies have linked Discord to harmful communities and cybercrime-related activities, including malware distribution, credential theft, phishing, doxxing, and sharing hacking tools \citep{heslep-berge-2024-mapping,sophos-2021-discord-malware,zscaler-2021-discord-cdn,intel471-2024-discord-cybercrime}. At the same time, communication in Discord communities often relies on shared norms, insider knowledge, and community-specific expressions \citep{nguyen-p-rose-2011-language,aquino-etal-2025-discord}. These properties make Discord a useful setting for studying why harmful messages can be difficult to interpret. This study therefore focuses on cybercrime-related Discord chats and examines how semi-closed communities, interleaved conversations, and community-specific expressions make harmful messages difficult to interpret.

\subsection{Automatic Analysis of Harmful Online Conversations}

A large body of work has studied automatic methods for detecting and analyzing harmful online conversations, including grooming, cyberbullying, aggression, harmful keyword extraction, and cyber threat intelligence extraction \citep{murnion-etal-2018-machine,vogt-etal-2021-early,alves-etal-2023-detecting,park-lee-2022-full}. These studies show that linguistic features and conversational context provide valuable signals for identifying harmful behavior.

However, harmful intent is often conveyed implicitly rather than explicitly. Prior work on implicit hate speech has shown that harmful intent may be expressed through coded language, euphemisms, sarcasm, and context-dependent implications \citep{elsherief-etal-2021-latent,wiegand-etal-2021-implicitly-abusive}. Cybercrime communities similarly use slang, abbreviations, dark jargon, and obfuscated expressions that can conceal meaning from outsiders while signaling trust or group membership to insiders \citep{magu-luo-2018-determining,yuan-etal-2018-reading,hughes-etal-2023-argot}.

Although prior work has examined linguistic challenges in harmful-content analysis, most studies have focused on detecting harmful behavior rather than explaining why certain messages are difficult to interpret. This study addresses this gap by constructing expert-reviewed interpretations of cybercrime-related Discord messages and using qualitative error analysis to identify recurring sources of interpretation difficulty.

\subsection{LLMs for Harmful Text Analysis}

Recent work has applied LLMs to harmful-content analysis tasks such as hate speech detection, content moderation, explanation generation, and annotation support \citep{ziems-etal-2024-large,guo-etal-2023-investigation,huang-etal-2023-chatgpt}. Because LLMs can leverage conversational context and broad linguistic knowledge, they may help interpret informal or context-dependent expressions that are not directly explained in the local text.

At the same time, LLM judgments are not always reliable. Prior studies have shown that LLM outputs can be affected by prompt design, contextual framing, cultural assumptions, and annotator perspectives \citep{masud-etal-2024-hate}. In implicit hate speech detection, LLMs may also overgeneralize harmful intent or misinterpret context-dependent expressions \citep{zhang-etal-2024-dont-go}. These limitations suggest that evaluating harmful-content analysis systems requires considering not only predictive accuracy but also the validity and interpretability of model explanations \citep{huang-2025-content}.

Despite growing interest in LLM-based harmful-content analysis, little is known about how humans and LLMs differ in interpreting harmful messages. Such differences matter because humans and LLMs may rely on different cues and fail in different ways. Understanding these differences can clarify which interpretation difficulties can be addressed by LLMs and which require external knowledge, human judgment, or additional contextual support. This study addresses this gap by analyzing interpretation difficulty in harmful Discord conversations and comparing human and LLM interpretations under varying contextual conditions.

\section{Experiments}

\subsection{Dataset Construction}

We use Discord messages collected in October 2023 by \citet{kawaguchi-etal-2024-chatgpt}. The original corpus contains 1,280,548 messages from 9,298 channels across 55 servers. The servers were collected using 196 keywords related to cybercrime and harmful activities, such as \textit{``drug''}, \textit{``malicious hacking''} and \textit{``scam''}.

We first excluded channels with fewer than 100 messages, because such channels contained too little conversational context for interpretation analysis. We then manually removed channels that consisted only of one-way advertising or promotional posts and did not contain meaningful conversational exchanges. From the remaining data, one author purposefully selected 100 target messages that appeared difficult to interpret due to slang, coded language, abbreviations, euphemisms, ambiguous expressions, or community-specific references. Because our goal is to characterize sources of interpretation difficulty rather than estimate their prevalence, we do not use random sampling.

The final dataset consists of 100 target messages. Although this is a relatively small sample, the size was chosen to enable detailed human annotation, discussion-based interpretation construction, expert review, confidence assessment, and subsequent evaluation for each message. This design reflects the exploratory goal of the study, which is to closely analyze sources of interpretation difficulty rather than to estimate their prevalence in the original corpus. To characterize the selected messages, we consulted WordNet 3.1 \citep{princeton-university-2010-wordnet} and Wiktionary \citep{wikimedia-foundation-2026-wiktionary}. Among the 100 messages, 92 contained terms whose relevant meanings were not registered in WordNet, and 73 contained terms whose relevant meanings were not registered in Wiktionary.

\subsection{Human Interpretation and Reference Construction}
\label{sec:human-reference}

Three graduate students with training in natural language processing interpreted each target message under three conditions with increasing access to contextual and external information. The interpretation was conducted under three conditions in a fixed order from A to C. Under each condition, annotators recorded a free-text interpretation of the target message as a natural-language statement. The interpretation was intended to paraphrase the likely meaning of the target message, including implied actions, referents, or domain-specific meanings when they could be inferred. To prevent information leakage across conditions, annotators recorded their interpretation under each condition before proceeding to the next one and were not allowed to revise earlier interpretations.

\begin{itemize}
\item \textbf{Condition A: Message only.}\ Annotators interpreted the target message without additional context.
\item \textbf{Condition B: Local context.}\ Annotators interpreted the target message with the 20 preceding and 20 following messages.
\item \textbf{Condition C: External knowledge.}\ Annotators interpreted the target message using the full channel history and external resources, including Web searches, online dictionaries, and local LLM assistance. In this condition, annotators also reported a confidence score from 0 to 100 for each interpretation, indicating how confident they were in the interpretation based on the available evidence.
\end{itemize}

After all annotators had completed all three conditions independently, they discussed their interpretations and established a single consensus interpretation for each message. During this discussion, annotators considered not only the content of the interpretation but also the degree of remaining uncertainty. The resulting consensus interpretations were assigned confidence scores, reviewed by an information security expert, and used as reference interpretations. These confidence scores were used to examine cases where the intended meaning remained uncertain even after consultation of extended context and external resources.

\subsection{LLM-based Interpretation}
\label{sec:llm-interpretation}

We evaluate two open-weight LLMs, GPT-OSS-20B and GPT-OSS-120B \citep{openai-gpt-oss}. We used open-weight models for security and privacy reasons, because the dataset contains sensitive harmful-content examples and should not be sent to external LLM providers. These models were selected because, at the time of our experiments, they represented some of the most recent open-weight models with strong general-purpose capabilities.

For generation, we used temperature $=1.0$ and top-$p=1.0$ for both models. We did not explicitly fix a random seed, so exact reproduction of individual generations may not be guaranteed. We did not tune prompts or decoding parameters because our goal was to evaluate baseline interpretive behavior rather than optimize performance.

For each target message, we generated interpretations under two conditions.

\begin{itemize}
\item \textbf{Condition i: Message only.}\\ The model received only the target message.
\item \textbf{Condition ii: Local context.}\\ The model received the target message with the 20 preceding and 20 following messages.
\end{itemize}

The prompts are shown in \appref{app:llm-prompts}. The models were instructed to provide a concise interpretation of the target message. No domain-specific persona, task-specific hints, retrieval augmentation, or examples were provided.

\subsection{Interpretation Evaluation}
\label{sec:interpretation-evaluation}

We recruited three graduate students who did not participate in reference interpretation construction. The evaluation targets were the human interpretations from Section~\ref{sec:human-reference} and the LLM interpretations from Section~\ref{sec:llm-interpretation}, each written as a free-text statement. Evaluators compared each candidate interpretation with the corresponding reference interpretation and assigned one of three labels: \textit{Match}, \textit{Partial Match}, or \textit{Mismatch}. The labels indicate whether the candidate interpretation conveyed the same intended meaning as the reference, captured it only partially, or failed to recover its core meaning. Detailed evaluation criteria and representative examples are provided in Appendix~\ref{app:evaluation-criteria}. Human and LLM interpretations were randomized before evaluation to reduce ordering effects and evaluator bias.

We used majority vote as the final evaluation label. To assess evaluator reliability, we calculated Fleiss' $\kappa$ \citep{fleiss1971measuring} and Krippendorff's $\alpha$ \citep{krippendorff2018content}. Both scores were 0.54, indicating moderate agreement. The confidence scores assigned during reference construction were not used to determine the evaluation labels, but only as an auxiliary signal for identifying reference interpretations with residual uncertainty.

Because Human Condition C allowed access to full channel history, external resources, and LLM assistance, it is not directly comparable to the LLM context condition. We therefore interpret Condition C primarily as evidence for the importance of external knowledge and extended context, rather than as a direct human--LLM capability comparison.

\begin{table}[t]
\centering
\caption{Interpretation results based on majority vote for 100 target messages. Human rows report the average counts across the three annotators. A and i indicate the message only condition; B and ii indicate the local context condition with the 20 preceding and 20 following messages; C indicates the condition with full channel history and external resources.}
\label{tab:majority-results}
\footnotesize
\setlength{\tabcolsep}{3pt}
\begin{tabular}{llrrr}
\toprule
Target & Cond. & Match & Partial & Mismatch \\
\midrule
Human avg. & A & 2.7 & 4.0 & 93.3 \\
Human avg. & B & 5.3 & 13.3 & 81.3 \\
Human avg. & C & 62.7 & 24.7 & 12.6 \\
\midrule
GPT-OSS-120B & i & 37 & 31 & 32 \\
GPT-OSS-120B & ii & 58 & 23 & 19 \\
GPT-OSS-20B & i & 33 & 37 & 30 \\
GPT-OSS-20B & ii & 41 & 32 & 27 \\
\bottomrule
\end{tabular}
\vspace{1mm}
\end{table}

\section{Results}
\label{sec:results}

Table~\ref{tab:majority-results} summarizes the results for human and LLM interpretations.

\subsection{Human Interpretations}

Human annotators had difficulty interpreting the target messages without additional information. In Condition A, where only the target message was shown, the average number of matches was 2.7 out of 100, while 93.3 messages were classified as mismatches. Providing local conversational context in Condition B led to only a small improvement, with the average number of matches increasing to 5.3 and mismatches decreasing to 81.3.

The largest improvement was observed in Condition C, where annotators could use the full channel history and external resources. In this condition, the average number of matches increased to 62.7, while mismatches decreased to 12.6. This suggests that local context alone is often insufficient, and that external knowledge and extended conversational context play an important role in interpreting difficult harmful messages.

Table~\ref{tab:condition-c-confidence} summarizes the confidence scores assigned to interpretations in Human Condition C. The scores were generally high, indicating that annotators were often able to reach plausible interpretations when extended context and external resources were available. However, some interpretations were assigned lower confidence scores, suggesting that the intended meaning remained uncertain for these messages. Such cases may include reference interpretations that do not fully recover the original intended meaning, and should therefore be interpreted with caution.

\begin{table}[t]
\centering
\caption{Confidence scores in Human Condition C by final evaluation label. Values indicate the average confidence score assigned by each annotator to their own interpretations, grouped according to whether the interpretation was later evaluated as Match, Partial Match, or Mismatch against the reference interpretation. The confidence scores did not show a clear difference according to agreement with the reference interpretation. The average confidence score reported during consensus construction of the reference interpretations was 92.1.}
\label{tab:condition-c-confidence}
\small
\setlength{\tabcolsep}{3pt}
\begin{tabular}{lrrr}
\toprule
Target & Match & Partial & Mismatch \\
\midrule
Annotator 1 & 79.3 & 80.0 & 80.0 \\
Annotator 2 & 75.3 & 65.0 & 79.4 \\
Annotator 3 & 80.9 & 79.4 & 68.1 \\
Annotator Average & 78.5 & 71.7 & 75.8 \\
\bottomrule
\end{tabular}
\end{table}

\subsection{LLM Interpretations}

Providing local context also improved LLM interpretations. GPT-OSS-120B increased from 37 matches in the message-only condition to 58 matches in the context condition. GPT-OSS-20B increased from 33 to 41 matches. For both models, the number of mismatches decreased when local context was provided.

GPT-OSS-120B outperformed GPT-OSS-20B under both conditions. In the context condition, GPT-OSS-120B achieved 58 matches, 23 partial matches, and 19 mismatches, which was the best result among the evaluated LLM settings.

\subsection{Comparison Between Human and LLM Interpretations}

Human Condition B and the LLM context condition used the same local context window, consisting of the target message with the 20 preceding and 20 following messages. Under this comparable input setting, the LLMs achieved substantially more matches than human annotators. This result suggests that the evaluated LLMs could often produce plausible interpretations from limited local context. However, the difference should not be attributed only to local context use, because LLMs may also rely on broad linguistic, cultural, and domain knowledge acquired during pretraining.

At the same time, human performance improved substantially in Condition C, where annotators could use full channel history, external resources, and LLM assistance. This condition is not directly comparable to the LLM context condition, which used only local context as input. We therefore interpret the improvement in Condition C as evidence for the importance of external knowledge and extended context, rather than as a direct measure of human superiority over LLMs.
\section{Error Analysis}

To better understand interpretation difficulty, we examined messages for which humans or LLMs produced incorrect interpretations. This analysis is exploratory and aims to identify recurring patterns associated with interpretation failures.

\subsection{Human Interpretation Errors}

Human errors were often associated with a lack of external knowledge. In Condition B, local context helped annotators understand the general topic of a conversation, but it was often insufficient for interpreting specific slang, coded terms, abbreviations, or community-specific expressions. Specifically, annotators could sometimes recognize that a conversation was related to cybercrime or illicit transactions, while still failing to identify the intended meaning of a particular term.

Performance improved substantially in Condition C, where annotators could use full channel history and external resources. Web searches and online dictionaries were useful for interpreting common slang or technical terms. However, some expressions remained difficult when they had multiple meanings or when their intended meanings were specific to a narrow community. For example, the message \textit{``Lemme google rw''} could be understood as a statement about searching for \textit{``rw''} but the intended meaning of \textit{``rw''} could not be determined from the available context or external resources.

Human errors were also caused by non-standard spelling and grammar. For example, the message \textit{``how do u loot ppl need to be a group memb''} contains informal spelling, omitted words, and unclear structure, which increased ambiguity and led to divergent interpretations.

\subsection{LLM Interpretation Errors}

LLM errors often occurred when coded expressions also had plausible literal meanings. In such cases, the models sometimes generated coherent interpretations based on the surface form of the message, but failed to recover the intended meaning in the community. For example, in the message \textit{``isn't that the cheese pizza video?''}, the model interpreted the phrase literally as a video about pizza rather than recognizing that, in this context, it was used as a coded reference to child sexual abuse material (CSAM).

Abbreviations and community-specific references were another source of LLM errors. For example, in the message \textit{``How much djs left?''}, \textit{``djs''} referred to David Jones, an Australian department store, but the model interpreted it as a misspelling or abbreviation related to discounts. This suggests that broad linguistic knowledge alone is not always sufficient when the intended meaning depends on local community knowledge.

At the same time, some surface-level obfuscations were less problematic for LLMs. For example, expressions such as \textit{``(C)lone C)ards)''} and \textit{``pu11ing m3th0d''} were sometimes interpreted correctly when sufficient context was available. Overall, LLM failures were most often associated with coded language, ambiguous abbreviations, multiple meanings, and community-specific knowledge that could not be recovered from local context alone.

\section{Classification of Interpretation Difficulty and Information Sources}

\begin{figure*}[t]
\centering
\includegraphics[width=\textwidth]{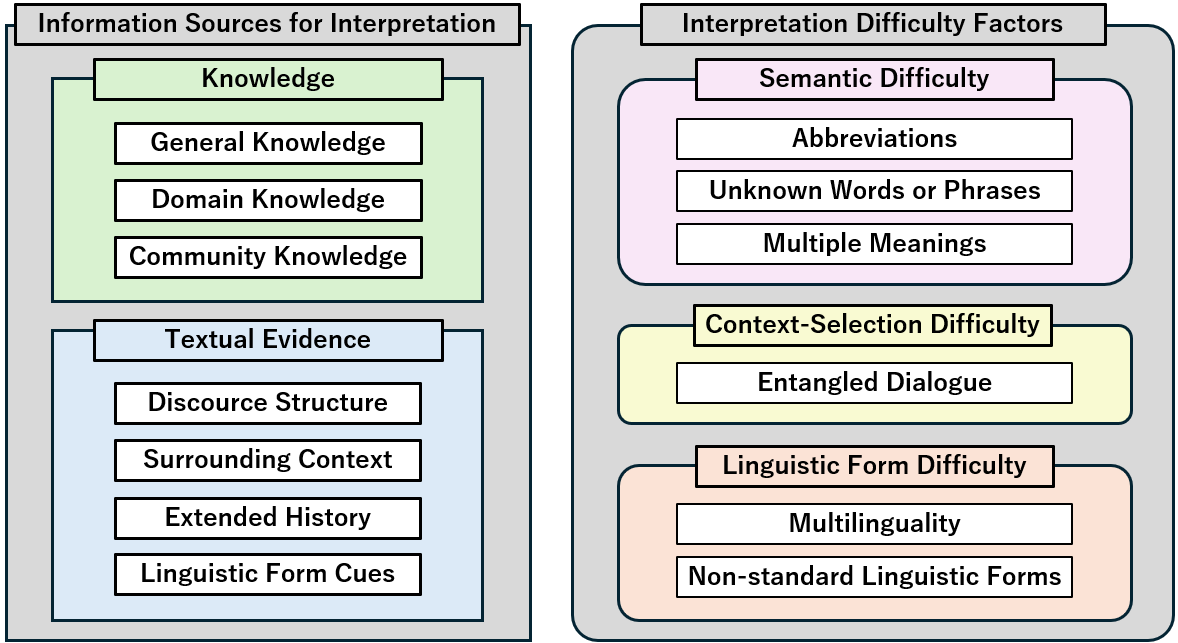}
\caption{Information sources and interpretation-difficulty factors in harmful online communication. The left panel summarizes sources of evidence that may support interpretation, while the right panel summarizes factors that make messages difficult to interpret. The figure does not assume a one-to-one mapping between information sources and difficulty factors.}
\label{fig:interpretation-difficulty-classification}
\end{figure*}

Based on the annotated messages, observations from the original Discord corpus, and prior literature, we organize interpretation difficulty using the classification shown in Figure 1. The classification separates two related aspects of interpretation: information sources that may support interpretation and factors that make messages difficult to interpret. This separation is important because a single difficulty factor does not correspond to a single required information source. For example, an abbreviation may be resolved through general knowledge, cybersecurity knowledge, community-specific knowledge, or extended conversational history. Conversely, one information source may help resolve several different difficulties. The purpose of this classification is therefore not to assign fixed solutions to each factor, but to clarify what kinds of evidence and difficulty are involved when humans and LLMs interpret harmful online communication.

\subsection{Information Sources for Interpretation}

We divide useful information sources into Knowledge and Textual Evidence.

\subsubsection{Knowledge}

Knowledge refers to information not necessarily contained in the immediate message.

\textbf{General Knowledge} includes common linguistic, cultural, and world knowledge, such as widely used slang, ordinary abbreviations, and background knowledge about online communication. LLMs may encode such knowledge through pretraining and can therefore sometimes interpret implicit meanings that are not obvious from the message alone \citep{brown-etal-2020-language,ziems-etal-2024-large}.

\textbf{Domain Knowledge} refers to knowledge about a specific activity domain. In this study, relevant domains include cybersecurity, cybercrime, illicit online markets, platform practices, and technical terminology. Such knowledge helps interpret messages that refer indirectly to malware, scams, credential theft, illicit trading, or other harmful activities. LLMs may possess partial domain knowledge, but their outputs can be unreliable when the required knowledge is long-tail, highly specialized, or rapidly changing \citep{kandpal2023large,mallen2023when}. This suggests that retrieval augmentation or expert-provided resources may be necessary when interpretation depends on information not reliably stored in model parameters \citep{lewis2020retrieval}.

\textbf{Community Knowledge} refers to knowledge specific to a particular group. This includes local conventions, repeated jokes, trust signals, shared assumptions, and community-specific uses of ordinary expressions. This type of knowledge is especially challenging for both humans and LLMs because it may not appear in public dictionaries, Web resources, or pretraining data. Even a capable LLM may generate a fluent but incorrect interpretation when a phrase has a local meaning that differs from its general or domain-level meaning. This problem is related to broader findings that implicit harmful meaning often depends on context, social assumptions, and community-specific implications rather than explicit lexical cues \citep{elsherief-etal-2021-latent,wiegand-etal-2021-implicitly-abusive,zhang-etal-2024-dont-go}.

\subsubsection{Textual Evidence}

Textual Evidence refers to information available in the message or conversation log.

\textbf{Discourse Structure} concerns how utterances are organized in multi-party conversation. In Discord-like environments, multiple dialogue threads may proceed simultaneously, so adjacent messages are not always relevant to the target message. Reply relations, speaker turns, time intervals, and topic continuity can help identify the correct thread. This problem is closely related to conversation disentanglement in multi-party chat settings \citep{elsner2008you,elsner2010disentangling,kummerfeld-etal-2019-large}. For LLMs, this creates a context-selection problem: even when relevant messages are included in the prompt, the model must determine which parts of the context should guide interpretation.

\textbf{Surrounding Context} refers to nearby messages around the target message. It may provide the immediate topic, addressee, or short-range references. In our experiments, the local-context condition used the 20 preceding and 20 following messages. This helped LLMs more than humans in many cases, but local context can still be insufficient when the relevant evidence is distant or sparse \citep{liu2023lost}.

\textbf{Extended History} refers to broader conversation history beyond the immediate window. It may include earlier uses of a term, recurring references, previous events, or background established long before the target message. Extended history is important for community-specific meanings, but it also raises challenges for LLM-based interpretation. Increasing context length does not automatically guarantee correct interpretation, because the model must still retrieve, prioritize, and integrate the relevant evidence from noisy conversation history \citep{liu2023lost,mallen2023when}.

\textbf{Linguistic Form Cues} include grammar, spelling, orthography, spacing, symbols, capitalization, and non-standard written forms. These cues help recover intended words and distinguish accidental errors from intentional distortion. Prior work on lexical normalization has shown that noisy online text requires recovering intended forms from non-standard spelling and informal writing \citep{han-baldwin-2011-lexical}. LLMs can often handle minor noise, but unusual tokenization, creative spelling, or obfuscation can still lead to incorrect interpretations, especially when surface form and intended meaning diverge \citep{chai2024tokenization}.

\subsection{Interpretation Difficulty Factors}

\subsubsection{Semantic Difficulty}

Semantic Difficulty arises when the meaning of a word, phrase, or expression is difficult to determine.

\textbf{Abbreviations} are difficult when their expansion is unclear or when several expansions are plausible. Some abbreviations are widely known, while others are domain-specific, platform-specific, community-specific, or temporary. Abbreviation expansion therefore requires contextual disambiguation, especially when multiple expansions are possible \citep{gorman2021structured}. LLMs may propose plausible expansions based on pretraining, but this can also produce overconfident errors when the local community uses an abbreviation differently from its common meaning.

\textbf{Unknown Words or Phrases} include emerging slang, technical terms, dark jargon, and localized expressions. Surrounding context may provide clues, but external or community-specific knowledge is often needed. For LLMs, these cases are difficult when the expression corresponds to rare or long-tail knowledge, or when the relevant meaning appeared after the model’s training data \citep{kandpal2023large,sun-etal-2024-toward,mei-etal-2024-slang}. Retrieval-based support may therefore improve interpretability by supplying up-to-date or domain-specific evidence \citep{lewis2020retrieval}.

\textbf{Multiple Meanings} create difficulty when an expression has both ordinary and specialized interpretations. This includes familiar phrases used as coded or community-specific references. Such cases are related to word sense disambiguation and to coded harmful language, where surface-level meaning may differ from intended social or harmful meaning \citep{navigli2009word,magu-luo-2018-determining,elsherief-etal-2021-latent}. LLMs are vulnerable in such cases because they can produce coherent literal interpretations while missing the coded meaning. This makes multiple meanings a key source of false-negative interpretation errors.

\subsubsection{Context-Selection Difficulty}

Context-Selection Difficulty arises when the main challenge is identifying which contextual information is relevant.

\textbf{Entangled Dialogue} occurs when the target message appears in a channel where multiple dialogue threads are interleaved. In such cases, interpretation fails not because the surrounding context is absent, but because the relevant thread is difficult to isolate from nearby irrelevant messages. For LLMs, larger context windows may introduce additional noise rather than resolve the ambiguity, unless the model can reliably track reply structure, speaker continuity, or topic flow \citep{elsner2008you,elsner2010disentangling,kummerfeld-etal-2019-large}.

\subsubsection{Linguistic Form Difficulty}

Linguistic Form Difficulty arises from how a message is written rather than from specialized meaning alone.

\textbf{Multilinguality} includes language differences, code-switching, and language-specific expressions. Online communities often include users with different linguistic backgrounds, and messages may mix languages or rely on culture-specific terms. LLM performance can vary across languages, especially in lower-resource or multilingual settings \citep{robinson2023chatgpt,huang2023not}. Multilinguality can therefore amplify both semantic and contextual uncertainty, particularly when a harmful or coded expression depends on language-specific slang or cultural knowledge.

\textbf{Non-standard Linguistic Forms} include informal grammar, typographical errors, spelling variation, leetspeak, and other non-standard spellings. These forms make interpretation difficult when the intended wording or sentence structure must first be recovered from noisy surface forms. Related challenges have been studied in lexical normalization of noisy online text \citep{han-baldwin-2011-lexical}. For LLMs, such cases may be easier when the distorted form remains close to a recognizable word, but errors can occur when tokenization or spelling variation prevents the model from recovering the intended expression \citep{chai2024tokenization}. Cases where an altered or ordinary-looking expression functions as a coded term are treated as Semantic Difficulty rather than Linguistic Form Difficulty.

\section{Conclusion}

This paper presented an exploratory study of interpretation difficulty in cybercrime-related Discord chats. Experiments with humans and LLMs showed that local context alone is often insufficient, while external knowledge and extended conversational history can substantially improve interpretation.

Building on these results, we organized interpretation difficulty along two aspects: the information needed to support interpretation and the factors that make messages difficult to understand, including semantic ambiguity, context selection, and non-standard linguistic forms.

These findings highlight that harmful chat analysis requires more than detecting harmful words. Reliable interpretation should instead be treated as an evidence-integration problem involving knowledge, conversational context, and robustness to informal or distorted language.

\setcounter{secnumdepth}{1}

\section*{Limitations}
This study has six limitations. First, our analysis focuses on text messages from Discord. Harmful online activity also occurs on other platforms, such as Telegram, Reddit, and social media, where communication styles and interpretation difficulties may differ \citep{la-morgia-etal-2021-uncovering,roy-etal-2025-darkgram,he-etal-2026-six-million}. In addition, we do not analyze images, videos, or audio, although these modalities are also important in harmful online activity \citep{ncmec-2025-cybertiplinedata}.

Second, the dataset consists of 100 messages that were purposefully selected because they appeared difficult to interpret. Therefore, the results should not be interpreted as prevalence estimates over the original Discord corpus or harmful online communication more broadly. The purpose of this study is to characterize sources of interpretation difficulty rather than measure how frequently they occur.

Third, the interpretation construction and evaluation were conducted in Japanese. This reduced differences in language ability among annotators and evaluators, but it may also affect the interpretation of English slang, coded terms, and culturally specific expressions. Future work should involve native speakers of the languages used in the chats and domain experts familiar with the relevant communities.

Fourth, although the reference interpretations were constructed through annotator discussion and expert review, they are not guaranteed to recover the original intended meaning in all cases. In particular, low-confidence reference interpretations indicate cases where ambiguity remained even after consulting extended context and external resources. Evaluation results involving such messages should therefore be interpreted with caution.

Fifth, the LLM experiments used two open-weight models and did not include retrieval augmentation or optimized prompting. Evaluating a wider range of models and information-access conditions would provide a more complete understanding of LLM interpretation ability.

Finally, the proposed classification was derived through post-hoc qualitative analysis. It organizes recurring sources of difficulty observed in our data, but further validation on independent datasets is necessary.

\section*{Ethics Statement}
The authors considered that this study did not require formal ethics review. Nevertheless, because this study analyzes sensitive harmful online communication, we describe the measures taken to reduce privacy, safety, and annotator well-being risks.

\subsection{Data Collection and Privacy Protection}

The data used in this study consist of messages collected from Discord servers that required invitation links \citep{kawaguchi-etal-2024-chatgpt}. Although these servers were not fully public, their invitation links were collected through publicly available server search sites. We therefore judged that users' expectations of privacy were lower than in completely closed communities. Nevertheless, because the data contain sensitive and potentially harmful content, we took measures to reduce privacy risks.

We do not analyze images, audio, or video, which are more likely to contain personally identifiable information. All examples presented in the paper were processed to prevent the identification of individuals or specific communities while preserving the meaning relevant to the analysis. During dataset construction and evaluation, personal and community-identifying information was removed or masked. The data are stored on a secure local server isolated from external access and are not made publicly available.

\subsection{Public Interest}

Because cybercrime and harmful online activity often occur in semi-closed communities, real-world data are important for understanding such communication and improving harmful-content analysis. We recognize the privacy and safety risks of working with such data and limit the analysis accordingly.

\subsection{Annotation and Evaluation}

The study involved human annotation and evaluation of harmful online messages. All annotators and evaluators completed data-handling training before the task. Because exposure to harmful content can cause psychological burden, participants were allowed to take breaks and stop the task if they felt distressed.

\subsection{Use of Local LLMs}

We used local LLMs for interpretation experiments to avoid sending sensitive research data to external providers. This reduced risks such as provider-side storage, secondary use for model improvement, or unintended exposure through logs. Research data, prompts, and model outputs were processed locally and were not intentionally sent to external LLM services.

\section*{Acknowledgments}
This paper is based on results obtained from a project, JPNP24003, commissioned by the New Energy and Industrial Technology Development Organization (NEDO).

\bibliography{custom}

\appendix
\section{Annotation Instructions}
\label{app:annotation-instructions}

This appendix describes the annotation procedure used to construct the reference interpretations.

Three annotators independently analyzed 100 Discord messages containing potentially difficult-to-interpret expressions, including slang, coded language, abbreviations, and community-specific terminology. The primary objective of the annotation process was to construct reliable reference interpretations for subsequent evaluation of human and LLM interpretations. 

The annotation process consisted of three interpretation conditions followed by a consensus step.

\subsection{Condition A: Message Only}

Annotators were shown only the target message.

They were instructed to describe the meaning of the message if they believed that the meaning could be inferred with reasonable confidence. Annotators were informed that perfect accuracy was not required and that tentative interpretations were acceptable. If the meaning could not be inferred, they were allowed to answer ``unknown.''

\subsection{Condition B: Local Context}

Annotators were shown the target message together with the 20 preceding and 20 following messages.

Annotators were asked to interpret the target message using the available conversational context. Annotators were informed that perfect accuracy was not required and that tentative interpretations were acceptable. If the meaning could not be inferred, they were allowed to answer ``unknown.''

\subsection{Condition C: External Knowledge}

Annotators were allowed to access the full channel history and consult external resources, including Web search engines, online dictionaries, and local LLMs.

Annotators were instructed to determine the most plausible interpretation based on all available evidence and to report a confidence score ranging from 0 to 100. Although LLMs could be used as supporting tools, annotators were required to make the final interpretation decision themselves.

\subsection{Consensus Interpretation}

After completing Conditions A to C independently, the three annotators discussed cases where their interpretations differed and established a single consensus interpretation for each message.

To prevent information leakage across conditions, annotators were not allowed to revise earlier interpretations after progressing to later stages. Thus, Conditions A to C were completed independently before any discussion occurred.

The resulting consensus interpretations were subsequently reviewed by an information security expert and used as the reference interpretations in the experiments.

\section{Prompts for LLM-based Interpretation Experiments}
\label{app:llm-prompts}

We intentionally avoided providing domain-specific personas (e.g., cybersecurity analyst, law enforcement investigator, or online safety expert). This design was motivated by two considerations. First, our goal was to evaluate the models' general interpretation capabilities rather than their performance under specialized prompting strategies. Second, assigning a security-related persona could bias the model toward interpreting ambiguous expressions as cybercrime- or harm-related content, potentially increasing false positives and over-attributing malicious intent. To reduce such bias, the models were prompted as general-purpose assistants and were asked only to interpret the meaning of the target message.

\subsection{System Prompt}

\begin{promptbox}
You are an assistant that explains the meaning of English text in natural Japanese.

Interpret the original intent and nuance as naturally as possible. Do not repeat the input sentence. Keep your answer concise and brief.
\end{promptbox}

\subsection{User Prompt}

\paragraph{Condition i: Message-only}

For the message-only condition, the target message was presented in isolation.

\begin{promptbox}
Please explain the meaning of the sentence surrounded by \(\star\) symbols. Provide a concise interpretation in a single sentence.

Example input:

\(\star\)i'd rug pull it at a lot lower haha\(\star\)
\end{promptbox}

\paragraph{Condition ii: Contextual}

For the context condition, the target message was presented together with the 20 preceding and 20 following messages.

\begin{promptbox}

Please explain the meaning of the sentence surrounded by \(\star\) symbols. Provide a concise interpretation in a single sentence.

Context:

[Previous 20 messages]

...

\(\star\)i'd rug pull it at a lot lower haha\(\star\)

...

[Following 20 messages]

\end{promptbox}

All model outputs were generated using the same prompts. No additional hints, examples, retrieval augmentation, or task-specific instructions were provided.

\begin{table*}[t]
\centering
\small
\caption{Detailed human interpretation results for each annotator (N=100).}
\label{tab:individual-human-results}
\begin{tabular}{llrrrr}
\toprule
Annotator & Condition & Match & Partial & Mismatch & Avg. Score \\
\midrule
Annotator 1 & A: message only & 2 & 3 & 95 & 2.93 \\
Annotator 1 & B: context & 7 & 15 & 78 & 2.71 \\
Annotator 1 & C: context + external knowledge & 74 & 20 & 6 & 1.32 \\
\midrule
Annotator 2 & A: message only & 2 & 8 & 90 & 2.88 \\
Annotator 2 & B: context & 3 & 21 & 76 & 2.73 \\
Annotator 2 & C: context + external knowledge & 56 & 28 & 16 & 1.60 \\
\midrule
Annotator 3 & A: message only & 4 & 1 & 95 & 2.91 \\
Annotator 3 & B: context & 6 & 4 & 90 & 2.84 \\
Annotator 3 & C: context + external knowledge & 58 & 26 & 16 & 1.58 \\
\bottomrule
\end{tabular}
\end{table*}

\section{Detailed Evaluation Criteria and Representative Examples}
\label{app:evaluation-criteria}

Three evaluators compared candidate interpretations against reference interpretations and assigned one of the following labels. Additional examples were provided to evaluators before the evaluation task. Evaluators were instructed to focus on semantic equivalence rather than surface-level wording similarity. One example of each label is shown below.

\subsection{Match (1)}

The candidate interpretation conveys essentially the same meaning as the reference interpretation. Minor differences in wording, level of detail, or phrasing are acceptable as long as the core meaning is preserved.

\begin{examplebox}
\textbf{Reference interpretation}

Asking PersonX to recommend an indica cannabis strain suitable for relaxation.

\textbf{Candidate interpretation}

Requesting that PersonX suggest an indica-type cannabis strain appropriate for relaxing.

\textbf{Reason}

Although the wording differs slightly, the core meaning is preserved.
\end{examplebox}

\subsection{Partial Match (2)}

The candidate interpretation captures part of the intended meaning but omits important information, contains minor inaccuracies, or introduces unsupported assumptions.

\begin{examplebox}
\textbf{Reference interpretation}

Effective pump-and-dump manipulation requires more than 1,000 participants and several large investors.

\textbf{Candidate interpretation}

Successfully increasing the price requires more than 1,000 participants and several whales.

\textbf{Reason}

The candidate correctly interprets the overall topic but fails to explain that ``whales'' refers to large investors.
\end{examplebox}

\subsection{Mismatch (3)}

The candidate interpretation substantially differs from the reference interpretation or fails to capture the core meaning.

\begin{examplebox}
\textbf{Reference interpretation}

The speaker tells their friends that Friday night has arrived and encourages them to wake up, smoke cannabis, and start the weekend in the best possible way.

\textbf{Candidate interpretation}

The speaker is casually and energetically telling their friends, “Friday is finally here! Wake up, get moving, and enjoy the weekend to the fullest.”

\textbf{Reason}

Although the candidate captures the general excitement about Friday and the weekend, it omits the key interpretation that the message encourages smoking cannabis. Because this important action is missing, the candidate fails to recover the core meaning of the reference interpretation.
\end{examplebox}

\section{Detailed Results for Individual Annotators}
\label{app:detailed-results}

This appendix provides detailed results for individual human annotators and reports inter-rater agreement for the evaluation task.

Table~\ref{tab:individual-human-results} presents the interpretation results of each annotator under Conditions from A to C. For descriptive analysis, we additionally report an average score, where Match, Partial Match, and Mismatch were assigned values of 1, 2, and 3, respectively. Lower scores therefore indicate better interpretation quality. While Table~\ref{tab:majority-results} in the main text reports averages across annotators, the table above shows the performance of each individual annotator. Consistent with the findings reported in Section~\ref{sec:results}, all annotators showed substantial improvement when external knowledge was available (Condition C), whereas only limited improvements were observed when local conversational context alone was provided (Condition B).

To assess the reliability of the evaluation process, we calculated inter-rater agreement among the three evaluators who compared candidate interpretations against the reference interpretations. Table~\ref{tab:inter-rater-agreement} reports Fleiss' $\kappa$ and Krippendorff's $\alpha$, both of which are widely used measures of agreement among multiple raters \citep{fleiss1971measuring,krippendorff2018content}.

The resulting values indicate moderate agreement among evaluators. Given that the task requires assessing semantic equivalence between free-form natural language interpretations rather than assigning predefined categorical labels, some degree of disagreement is expected. Nevertheless, the observed agreement suggests that the evaluation criteria provided in \appref{app:evaluation-criteria} enabled reasonably consistent judgments across evaluators.

\begin{table}[t]
\centering
\small
\caption{Inter-rater agreement in the evaluation task.}
\label{tab:inter-rater-agreement}
\begin{tabular}{lrl}
\toprule
Metric & Value & Interpretation \\
\midrule
Fleiss' $\kappa$ & 0.540 & Moderate \\
Krippendorff's $\alpha$ & 0.540 & Moderate \\
Exact agreement rate & 0.616 & -- \\
\bottomrule
\end{tabular}
\end{table}

\end{document}